\documentclass[10pt,twocolumn,letterpaper]{article}

\newif\ifarxiv
\arxivtrue

\ifarxiv
  \usepackage[pagenumbers]{cvpr}
\else
  \usepackage[review]{cvpr}
\fi

\usepackage{tabularx} 
\usepackage{booktabs} 
\usepackage{makecell} 
\usepackage{multirow} 
\usepackage{colortbl}
\usepackage{graphicx}
\usepackage{pifont} 
\usepackage{xcolor} 
\usepackage{textcomp}
\usepackage{gensymb}
\usepackage{enumitem} 

\usepackage{microtype}
\AtBeginDocument{\parfillskip=0pt plus 0.8\columnwidth}

\usepackage[dvipsnames]{xcolor}

\newcommand{\ours}{CalfVO\xspace}
\newcommand{\boldparagraph}[1]{\vspace{2pt}\noindent{\bf #1}}

\colorlet{colorFst}{Green!25}       
\colorlet{colorSnd}{SpringGreen!45} 
\colorlet{colorTrd}{Yellow!30}      
\colorlet{colorLow}{black!45}       
\newcommand{\lo}{\color{colorLow}}   

\definecolor{cvprblue}{rgb}{0.21,0.49,0.74}
\usepackage[pagebackref,breaklinks,colorlinks,citecolor=cvprblue]{hyperref}

\ifarxiv\else
  \usepackage{xr}
\fi

\def\paperID{459} 
\def\confName{3DV\xspace}
\def\confYear{2027\xspace}

\title{Monocular Visual Odometry without Calibration or Test-time Optimization}

\author{%
Vladimir Yugay\thanks{Equal contribution. Corresponding author:
\texttt{voviktyl@gmail.com}}\quad
Duy-Kien Nguyen\footnotemark[1]\quad
Theo Gevers\quad
Cees G. M. Snoek\quad
Martin R. Oswald\\[2pt]
University of Amsterdam\\
{\small\url{https://vladimiryugay.github.io/calfvo}}%
}

\begin{document}
\renewcommand{\thefootnote}{\fnsymbol{footnote}}
\maketitle
\renewcommand{\thefootnote}{\arabic{footnote}}\setcounter{footnote}{0}
\begin{abstract}
The most accurate monocular visual odometry systems require known camera intrinsics, refine their
estimates with test-time optimization, and recover trajectories only up to an unknown factor.
Systems built on large 3D models need no intrinsics, but they remain considerably less accurate and slower for odometry.
Direct pose regression avoids all these requirements, yet it has not matched either approach's accuracy.
We revisit this formulation with a transformer that predicts
relative camera poses together with separate rotation and translation confidences over overlapping
image windows, supervised by camera poses alone. A confidence-weighted module then aggregates the
overlapping predictions into a single trajectory. The resulting method, \ours, needs no intrinsics,
no bundle adjustment, and no loop closure, and it recovers scale from learned priors,
accurately enough that it is evaluated without any alignment to the ground truth. Across five
benchmarks, it is the most accurate calibration-free method on every metric we report, and it runs at
53 FPS, faster than every baseline.
\end{abstract}

%
\begin{figure}[t]
    \centering
    \includegraphics[width=0.82\linewidth]{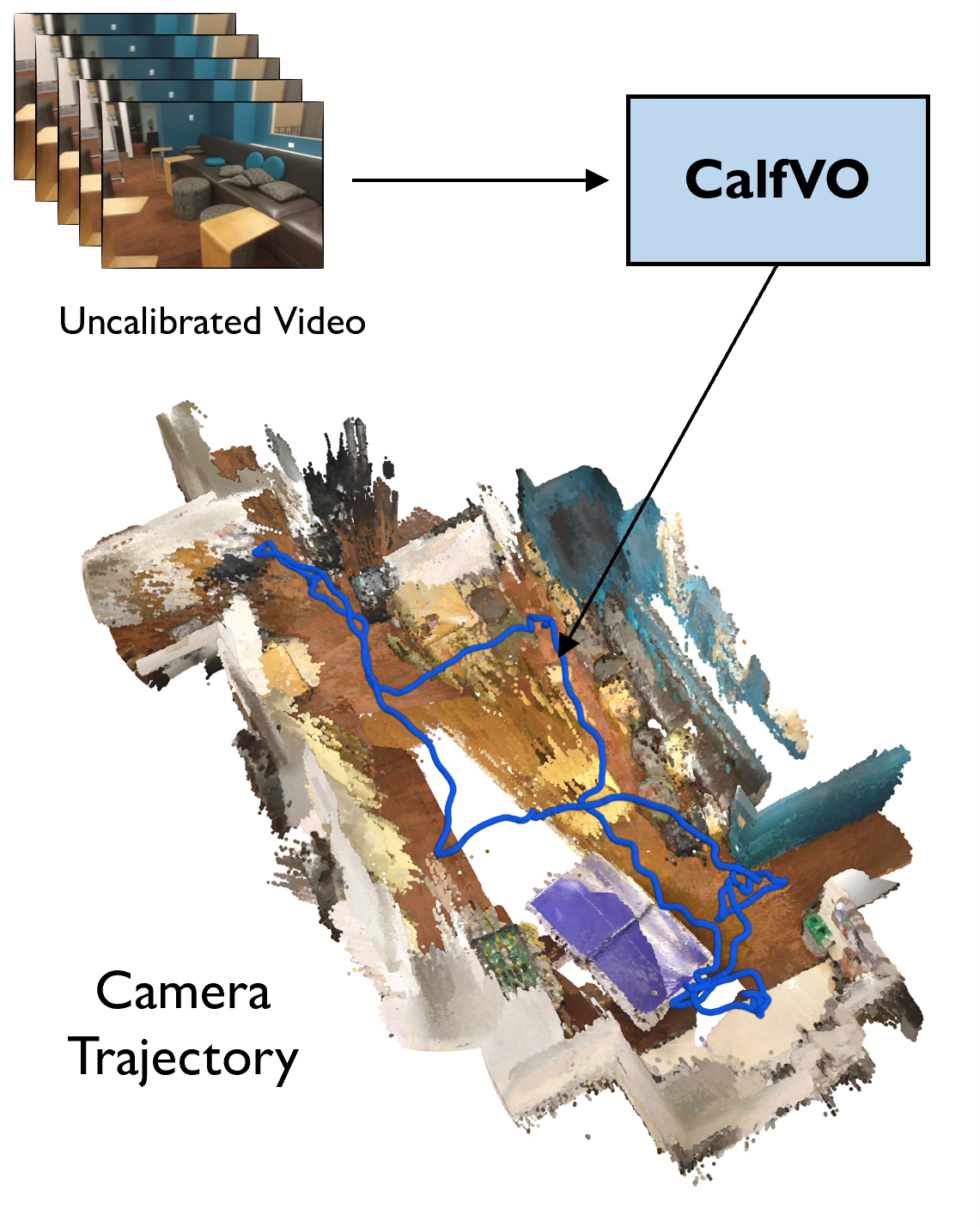}
    \caption{\textbf{Visual odometry without calibration or alignment.} Given an uncalibrated video, \ours predicts a metric camera trajectory in a single forward pass at \emph{53 FPS}, with no camera intrinsics, no bundle adjustment, and no loop closure. The trajectory is placed in the scene at the scale the network outputs, with no fitting to ground truth. The scene is rendered from ground-truth depth and is not predicted by our method.}
    \label{fig:teaser}
\end{figure}
\section{Introduction}
\label{sec:intro}
The goal of visual odometry (VO) is to estimate a camera's position and orientation from a sequence of
video frames~\cite{cadena2016past_present_future_of_slam}, with applications in augmented and virtual
reality, autonomous driving and robotics. Monocular VO is the harder setting. Stereo
vision~\cite{wang2017stereo_dso,7353631} and additional sensing such as inertial
measurements~\cite{stumberg2018direct_sparse_visual_inertial,christian2015robotics_science_systems}
infer the scale of the motion and supply geometric constraints that one camera cannot. A single
camera, though, requires no additional hardware, which is why we address the monocular case here.

That advantage, however, is only partly realized in practice. Classical systems and modern
learning-based ones alike~\cite{chen2024leapvo,teed2023dpvo} recover motion by enforcing geometric
constraints between frames. These constraints are written in terms of the camera intrinsics, which are
unavailable for crowdsourced footage, internet video, or devices whose optics change between
recordings, and estimating them leaves a residual error that propagates into the trajectory. The
constraints also leave the scale of the motion undetermined, so the trajectory is recovered only up to
an unknown factor. Finally, the hand-crafted components that enforce them limit what the learned part
of these pipelines can absorb from large and diverse training data.

Large 3D models~\cite{wang2024dust3r,leroy2024mast3r,wang2025vggt} require no intrinsics, inferring
geometry and camera pose from pixels alone, but they are trained for sparse-view reconstruction under
dense 3D supervision and process only a limited number of frames at once, which leaves them
inapplicable to long video. Systems built on them~\cite{murai2024mast3rslam,maggio2025vggt-slam,wang2026vggtomega}
extend the formulation to long videos with optimization backends, yet they remain inaccurate for
odometry and slow.

Regressing camera pose directly from pixels avoids these requirements altogether, since it needs
neither intrinsics nor an optimization backend. TSformer~\cite{francani2025tsformer} showed that this
is feasible, but its accuracy remains far behind optimization-based pipelines, which has
kept the formulation out of practical use. We revisit this formulation with a transformer~\cite{vaswani2017transformer} that regresses pose directly from
video, which reduces the inductive bias of the pipeline and lets the model learn from large and
heterogeneous training data~\cite{nguyen2025pit}. Trained from pose supervision alone, it predicts a
separate confidence for rotation and for translation, which balances the two error terms in a way no
fixed loss weighting can. Aggregating the predictions of overlapping windows then extends the model to
videos of arbitrary length in place of a backend. Scale follows from the
learned priors rather than from a separate mechanism.

We introduce \textbf{Cal}ibration-\textbf{f}ree \textbf{V}isual \textbf{O}dometry (\ours), a pipeline
for monocular visual odometry that needs no camera intrinsics, no bundle adjustment and no loop
closure, and which recovers scale from learned priors rather than geometry, accurately enough to be
used without rescaling. We evaluate it on five
benchmarks that span indoor, driving and aerial motion, real and synthetic imagery, and different
camera models, and it is the most accurate calibration-free method on every metric we report. Our main contributions are:
\begin{itemize}
    \item An efficient odometry architecture that factorizes attention over time and space and
      regresses relative poses from a frozen encoder, running at 53 FPS.
    \item An uncertainty-aware training objective in which the model predicts separate rotation and
      translation log-variances alongside each pose, learned from pose supervision alone.
    \item An inference module that aggregates the predictions of overlapping windows into one
      trajectory, replacing the optimization backend.
\end{itemize}

\section{Related Work}
\label{sec:related_work}

\boldparagraph{Visual odometry.}
Visual odometry (VO) systems estimate a camera's trajectory from video. Unlike SLAM methods that mitigate error accumulation through loop closure~\cite{cadena2016past_present_future_of_slam,campos2021orbslam3,yugay2024magicslam}, VO operates without global correction and is therefore subject to drift. Additional sensing, as in visual--inertial~\cite{christian2015robotics_science_systems,stumberg2018direct_sparse_visual_inertial} and stereo~\cite{7353631,wang2017stereo_dso} odometry, improves accuracy at the cost of specialized hardware. Traditional monocular approaches~\cite{engel2014lsd,engel2018direct,campos2021orbslam3} instead struggle with ill-posed scale estimation, relying on hand-crafted constraints and on camera parameters that are sensitive to calibration. \ours uses a transformer to extract representations that encode implicit scale and motion priors. It resolves scale ambiguity from visual cues alone and needs no intrinsics.

\begin{figure*}[!t]
  \centering
  \includegraphics[width=\linewidth]{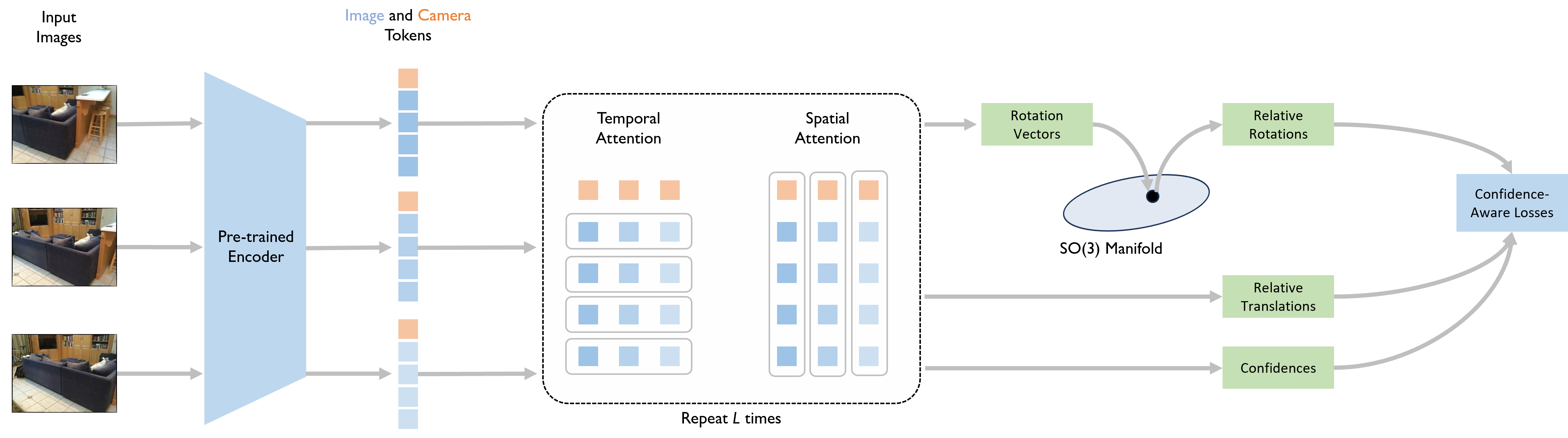}\\
  \caption{\textbf{Odometry transformer architecture.} Given multiple input frames, a frozen image encoder extracts per-image token embeddings. Camera embeddings are then concatenated to aggregate the information for camera pose estimation. The embeddings are decoded by $L$ repeating decoder blocks with temporal and spatial attention modules. The rotations are projected onto the $\mathbb{SO}(3)$ manifold to ensure valid relative rotations.}
  \label{fig:architecture}
\end{figure*}

\boldparagraph{Deep monocular visual odometry.} Deep learning has advanced monocular VO in both supervised~\cite{wang2017deepvo,tartanvo2020tartanVO,teed2020deepv2d,teed2023dpvo} and unsupervised~\cite{alisha2019unsupervised_learning_of_depth_and_ego_motion,li2020selfsuperviseddeepvisualodometry,wimbauer2025anycam} settings, evolving from early recurrent networks~\cite{wang2017deepvo} to recent geometric methods like DPVO~\cite{teed2023dpvo} and LeapVO~\cite{chen2024leapvo}. AnyCam~\cite{wimbauer2025anycam} also operates without intrinsics, but is trained self-supervised, consumes pre-trained depth and flow, and recovers trajectories only up to scale. Its trajectory refinement is a global bundle adjustment, which we found does not scale to the video lengths of our benchmarks, even with video subsampling. While these recent models achieve state-of-the-art accuracy through iterative updates and keypoint tracking, they inherit two requirements from classical pipelines, namely precise camera calibration and computationally expensive bundle adjustment. These dependencies, combined with the inability to recover absolute metric scale, restrict deployment in unconstrained settings.

Motivated by these limitations, direct regression methods like TSformer~\cite{francani2025tsformer} have been proposed. They dispense with both requirements, but remain well behind optimization-based pipelines in accuracy, and they integrate relative poses without suppressing unreliable ones. We attribute this to the architecture and to the absence of a robust inference mechanism. In contrast, \ours pairs a transformer with confidence-weighted aggregation over overlapping windows, regressing camera trajectories without intrinsics and without post-optimization, and recovering their scale accurately enough that no alignment to the ground truth is needed.

\boldparagraph{Reconstruction with 3D foundation models.} Transformers trained on large-scale data now serve as calibration-free foundation models for 3D geometry. DUSt3R~\cite{wang2024dust3r} and MASt3R~\cite{leroy2024mast3r}, built on cross-view completion pre-training~\cite{philippe2022croco}, infer point maps and camera pose directly from image pairs without known intrinsics. The approach has since been extended to multi-view sets and dynamic scenes~\cite{wang2025vggt,depthanything3,zhang2024monst3r}, and these models estimate camera pose without any camera knowledge. Trained for sparse-view reconstruction under dense 3D supervision, they attend jointly over all input views and process only a limited number of frames at once, which leaves them inapplicable to long videos. Memory-based variants~\cite{wang20243d_reconstruction_with_spatial_memory,cabon2025must3r,wang2025cut3r} relax this by accumulating a state as frames arrive. To process long videos, recent systems instead pair the foundation models with optimization backends, such as bundle adjustment or manifold alignment~\cite{murai2024mast3rslam,maggio2025vggt-slam,wang2025amb3r,wang2026vggtomega}. The backends dominate the inference time of these pipelines, which remain less accurate for odometry than methods built for it. \ours instead processes videos of arbitrary length without a backend, aggregating overlapping window predictions in a single pass.
\section{Method}
\label{sec:method}
Given a monocular video sequence $V \in \mathbb{R}^{N \times H \times W \times 3}$ consisting of $N$ frames of height $H$ and width $W$, our objective is to estimate the camera's trajectory over time, including its scale. We represent the trajectory as a sequence of camera poses $\{\mathbf{T}_i\}_{i=1}^{N}$, where each pose $\mathbf{T}_i \in \mathbb{SE}(3)$ describes the camera position and orientation at frame $i$. Our model predicts relative camera poses between pairs of input frames along with a per-prediction uncertainty, which an inference module then aggregates into a global trajectory. A high-level overview of the proposed architecture is shown in~\cref{fig:architecture}.
\subsection{Architecture}
\boldparagraph{Encoder.} We adopt a frozen ViT-Large~\cite{dosovitskiy2020vit} encoder following the CroCo~\cite{philippe2022croco} architecture and trained within the DUSt3R~\cite{wang2024dust3r} framework. Each frame is split into $(h \cdot w)$ non-overlapping patches of size $p$, with $h = H/p$ and $w = W/p$, which the encoder maps to features $F \in \mathbb{R}^{K \times (h \cdot w) \times d}$ for each window of $K$ frames. We compare backbone choices in \cref{tab:ablations}.

\boldparagraph{Time--space decoder.} The decoder is a stack of $L$ identical layers, each applying multi-head temporal attention, then multi-head spatial attention, then a feed-forward network, factorizing attention over the two axes as in video transformers~\cite{zhang2021vidtr}.
To summarize the information relevant to camera pose prediction, we introduce a single learnable camera token that propagates across the spatial attention layers. The camera token participates only in spatial attention, while temporal attention operates on the image tokens alone.

Formally, let $X_n \in \mathbb{R}^{K \times (h \cdot w) \times d}$ denote the image tokens at the input of the $(n+1)^\text{th}$ decoder layer and $\mathbf{q}_n \in \mathbb{R}^{d}$ the accompanying camera token, with $X_0 = F + E_{\text{pos}} + E_{\text{time}}$ for learned positional and temporal embeddings, and $\mathbf{q}_0$ a learned embedding. Writing $\mathrm{MHA}$ for standard multi-head scaled dot-product attention~\cite{vaswani2017transformer} applied along a given axis and $\mathrm{LN}$ for layer normalization~\cite{ba2016layernormalization}, one decoder layer computes
\begin{align}
  \hat{X}_n &= X_n + W_t\,\mathrm{MHA}_{\text{time}}(\mathrm{LN}(X_n)), \\
  \bigl[\tilde{Q}_n, \tilde{X}_n\bigr] &= Z_n + \mathrm{MHA}_{\text{space}}(\mathrm{LN}(Z_n)), \\
  \bigl[\mathbf{q}_{n+1}, X_{n+1}\bigr] &= Y_n + \mathrm{FFN}(\mathrm{LN}(Y_n)),
\end{align}
with $Z_n = [\mathbf{1}_K \mathbf{q}_n, \hat{X}_n]$ the image tokens carrying the broadcast camera
token, and $Y_n = [\tfrac{1}{K}\sum_{i} \tilde{Q}_{n,i},\, \tilde{X}_n]$ the same stack after the $K$
camera slots are averaged back into one. $\mathrm{MHA}_{\text{time}}$ attends across the $K$ frames independently at each of the $h \cdot w$ spatial positions and is followed by a learned projection $W_t$, while $\mathrm{MHA}_{\text{space}}$ attends jointly over the $h \cdot w$ image tokens and the camera token of each frame. Averaging the camera slots before the feed-forward network leaves one descriptor that accumulates evidence from the whole window. Each of the two attention modules has its own learned query, key, value, and output projections.

\subsection{Relative Camera Pose Regression}
Given a window of $K$ input frames, \ours predicts relative camera transformations and associated uncertainties between consecutive frames. The regression head reads the final camera token $\mathbf{q}_L$ together with $K$ per-frame descriptors, each formed by mean-pooling the patch tokens of that frame in $X_L$. These $1 + K$ descriptors are concatenated into a single vector of dimension $(1+K)d$, which one linear layer maps to $14(K-1)$ outputs. Reshaping gives one $14$-dimensional vector per consecutive frame pair, so a window of $K$ frames yields the $K-1$ relative poses along it. For each pair $(i, i+1)$, the $14$ values comprise an unconstrained rotation matrix $\tilde{\mathbf{R}}_{i,i+1} \in \mathbb{R}^{3 \times 3}$, a translation vector $\mathbf{t}_{i,i+1} \in \mathbb{R}^{3}$, and log-variance scalars $\mathbf{u}_R, \mathbf{u}_t \in \mathbb{R}$ for rotation and translation, respectively.

To ensure a valid rotation prediction, $\tilde{\mathbf{R}}_{i,i+1}$ is projected onto the $\mathbb{SO}(3)$ manifold using the special orthogonal Procrustes problem~\cite{bregier2021deepregression} by minimizing the Frobenius norm $\|\cdot\|_F$ of the matrix residual:
\begin{equation}
\hat{\mathbf{R}}_{i,i+1} =
\arg\min_{\mathbf{R} \in \mathbb{SO}(3)}
\|\mathbf{R} - \tilde{\mathbf{R}}_{i,i+1}\|_F^2.
\end{equation}
The solution is obtained via singular value decomposition as in~\cite{umeyama1991lse}.
\subsection{Uncertainty-Aware Pose Learning}
Reconstruction models learn a dense uncertainty, one value per point of the predicted point map, without requiring explicit labels~\cite{wang2024dust3r,wang2026vggtomega}. We instead place the uncertainty on the camera, adopting a heteroscedastic formulation~\cite{cipolla2018uncertainity} in which the network
predicts one log-variance for rotation and one for translation. Note that $\exp(-\mathbf{u})$ therefore acts as
a precision, so a larger $\mathbf{u}$ denotes a less reliable prediction. The rotation loss is a geodesic loss on $\mathrm{SO}(3)$ between the predicted rotation $\hat{\mathbf{R}} \in \mathbb{SO}(3)$ and the ground-truth rotation $\mathbf{R} \in \mathbb{SO}(3)$:
\begin{equation}
\mathcal{L}_{\text{rot}} =
\cos^{-1}
\left(
\frac{\mathrm{Tr}(\mathbf{R}^\top \hat{\mathbf{R}}) - 1}{2}
\right),
\end{equation}
and the translation error is defined as an L1 loss:
\begin{equation}
\mathcal{L}_{\text{trans}} =
\|\mathbf{t} - \hat{\mathbf{t}}\|_1.
\end{equation}
where $\hat{\mathbf{t}} \in \mathbb{R}^3$ and $\mathbf{t} \in \mathbb{R}^3$ are the predicted and ground-truth relative translations respectively. Both losses are optimized together using $\mathbf{u}_R$ and $\mathbf{u}_t$:
\begin{equation}
\begin{split}
\mathcal{L}
=~&
\mathcal{L}_{\text{rot}} \exp(-\mathbf{u}_R) + \mathbf{u}_R \\
+~&
\mathcal{L}_{\text{trans}} \exp(-\mathbf{u}_t) + \mathbf{u}_t.
\end{split}
\end{equation}
This formulation penalizes overconfident low-accuracy predictions through the additive $\mathbf{u}$ term while down-weighting uncertain residuals via $\exp(-\mathbf{u})$.
\subsection{Inference Module}
In visual odometry, a single wrong prediction can severely affect the entire trajectory because there is no global optimization. Therefore, we predict each relative pose from several overlapping windows and average them, weighted by the learned uncertainties. For window size $K$ and window advance $\Delta < K$, the input video on $N$ consecutive frames is decomposed into overlapping windows $\{1, \dots ,K\}, \{1 + \Delta, \dots ,K + \Delta\}, \cdots$, the last one ending at $N$. \ours predicts relative rotations, translations, and uncertainties for every window. Due to the overlap, the same relative pose is predicted multiple times from different contexts, yielding complementary estimates. Let
\[
\{(\mathbf{R}_{i,j}^{(k)}, \mathbf{t}_{i,j}^{(k)},
\mathbf{u}_R^{(k)}, \mathbf{u}_t^{(k)})\}_{k=1}^{M}
\]
denote multiple predictions of the relative transformation $(i,j)$. For $s \in \{R, t\}$, the predicted log-variances are converted into precisions and normalized into weights over the $M$ predictions,
\begin{equation}
\tilde{w}_s^{(k)} = \frac{\exp(-\mathbf{u}_s^{(k)})}{\sum_{\ell=1}^{M} \exp(-\mathbf{u}_s^{(\ell)})},
\end{equation}

The confidence-weighted average rotation is the weighted chordal $L_2$ mean on $\mathbb{SO}(3)$~\cite{hartley2013rotaveraging}:
\begin{equation}
\bar{\mathbf{R}}_{i,j} =
\arg\min_{\mathbf{R} \in \mathbb{SO}(3)}
\sum_{k=1}^{M} \tilde{w}_R^{(k)} \, \|\mathbf{R} - \mathbf{R}_{i,j}^{(k)}\|_F^2,
\end{equation}
which admits a closed-form solution as the leading eigenvector of the weighted outer-product matrix of the corresponding unit quaternions~\cite{markley2007quaternionaveraging}. The mean is taken over at most $\lceil (K-1)/\Delta \rceil$ rotations, three in our setting, so the aggregation requires no iterative optimization and adds a constant cost per edge. The confidence-weighted average translation is obtained via:
\begin{equation}
\bar{\mathbf{t}}_{i,j} = \sum_{k=1}^{M} \tilde{w}_t^{(k)} \mathbf{t}_{i,j}^{(k)}.
\end{equation}
The two are stacked into the fused transformation $\bar{\mathbf{T}}_{i,j} \in \mathbb{SE}(3)$ with $\bar{\mathbf{R}}_{i,j}$ as its rotation block and $\bar{\mathbf{t}}_{i,j}$ as its translation. After confidence-based averaging of duplicated edges, the global trajectory is obtained by sequential composition:
\begin{equation}
\mathbf{T}_0 = \mathbf{I},
\qquad
\mathbf{T}_{i+1} =
\mathbf{T}_i \bar{\mathbf{T}}_{i,i+1}.
\end{equation}
More details are provided in the supplementary.

\section{Experiments}
\label{sec:experiments}

\boldparagraph{Datasets.} We train on a mixture of four datasets and evaluate on five test sets at
every frame, taking the test frames consecutively and without temporal subsampling. We describe each
dataset in turn, in the order used in every table.
ScanNet~\cite{dai2017scannet} gives 1513 real indoor scenes for training and 97 of the 100 scenes of
its test split for evaluation, the remaining three were dropped because their camera pose annotations
contain invalid entries. KITTI~\cite{geiger2012kitti} adds roughly 20k frames of real driving data
under the splits of~\cite{8100183, Godard2018DiggingIS, 3454287.3454291}, which hold out sequences 09
and 10 for testing, and we take all 50 sequences of its synthetic counterpart, Virtual
KITTI~\cite{gaidon2016virtual} for training only.
TartanAir~\cite{wang2020tartanair} supplies most of the synthetic motion, 1088 of the 1122 sequences of
v2 for training with the SoulCity and Ocean environments held out in full, and 18 v1 sequences from
those two environments for evaluation, taken from the iSLAM~\cite{fu2024islam} test set. Finally,
TUM\_RGBD~\cite{sturm2012tumrgbd} and EuRoC~\cite{burri2016euroc} appear in training in no form and
serve for evaluation only, with 9 and 11 sequences following the split of
MASt3R-SLAM~\cite{murai2024mast3rslam}.

\boldparagraph{Evaluation Metrics}. Because \ours performs neither loop closure nor global
bundle adjustment, we follow MAC-VO~\cite{qiu2025mac} and quantify accuracy with relative rather
than global trajectory error. We report the relative translation error $t_{\text{rel}}$
(m/frame) and the relative rotation error $r_{\text{rel}}$ (\degree/frame),
\begin{align}
  t_{\text{rel}} &= \frac{1}{N-1}\sum_{t=1}^{N-1}
    \left\lVert \mathbf{p}_{t+1} - \mathbf{p}_{t}
      - R_{t}\hat{R}_{t}^{\top}\bigl(\hat{\mathbf{p}}_{t+1} - \hat{\mathbf{p}}_{t}\bigr)
    \right\rVert_{2}, \\
  r_{\text{rel}} &= \frac{180}{\pi}\,\frac{1}{N-1}\sum_{t=1}^{N-1}
    \left\lVert \log\bigl(\hat{R}_{t,t+1}^{\top} R_{t,t+1}\bigr) \right\rVert_{2},
\end{align}
where $\mathbf{p}_{t}$ and $R_{t}$ are the ground-truth position and rotation at frame $t$,
$\hat{\mathbf{p}}_{t}$ and $\hat{R}_{t}$ the corresponding estimates, and
$R_{t,t+1} = R_{t}^{\top}R_{t+1}$ the relative rotation between consecutive frames.
Where a baseline is up to scale, it is aligned to the ground truth with a $\mathrm{Sim}(3)$
transformation over rotation and scale before the metrics are computed, following the protocol under
which these methods report their results. The transformation is fitted per sequence, so those
baselines are handed a scale and a rotation that \ours has to predict, and the comparison is
conservative in their favor. In \cref{tab:tracking_accuracy}, $\textbf{best}$ and
$\underline{\text{second best}}$ results are highlighted.

\boldparagraph{Implementation details.} The encoder is a frozen
CroCoV2~\cite{philippe2022croco} ViT-Large trained within the DUSt3R~\cite{wang2024dust3r} framework,
roughly 300 million frozen parameters, patched with
flash-attention~\cite{dao2023flashattention2}. The decoder is 12 alternating
time--space attention blocks totaling 200 million parameters, and the model takes 8 input views
resized to $224 \times 224$. We train with the AdamW~\cite{loshchilov2017adamw} optimizer for 57k steps
at a batch size of 20. Each training window is drawn at a random frame stride, from 1 to 5 for the
indoor and synthetic sets and 1 to 2 for the two driving sets, since a vehicle advances far further
between consecutive frames than a handheld camera does. The model therefore sees a range of inter-frame
motions rather than a single frame rate. The inference module uses a window size of $K = 8$, matching
the number of input views, with consecutive windows overlapping by 5 frames.
Every number we report comes from the final checkpoint of a single run, with no per-dataset checkpoint
selection. The supplementary material reports the full training setup, the confidence ablation on all
five benchmarks, the overlap sweep, the accuracy of the estimated intrinsics, decoder
attention maps, sequence completion rates, and a failure-mode analysis.

\begin{table*}[!t]
\caption{Relative translation error $t_{\text{rel}}$ (m/frame) and relative rotation error $r_{\text{rel}}$ (\degree/frame). Benchmarks marked \textsuperscript{*} also contribute training data, with the test scenes, environments, or sequences held out. Rows marked \dag~are up to scale and are $\mathrm{Sim}(3)$-aligned before evaluation. Gray rows use the ground-truth intrinsics. The uncalibrated block holds the self-calibrated methods, which estimate the intrinsics they need with GeoCalib, above the calibration-free ones, which need none. Bold is best and underline is second best among the uncalibrated rows.}
\label{tab:tracking_accuracy}
\centering
\footnotesize
\setlength{\tabcolsep}{3pt}
\renewcommand{\arraystretch}{1.05}
\begin{tabularx}{\textwidth}{c|l|>{\centering\arraybackslash}X>{\centering\arraybackslash}X|>{\centering\arraybackslash}X>{\centering\arraybackslash}X|>{\centering\arraybackslash}X>{\centering\arraybackslash}X||>{\centering\arraybackslash}X>{\centering\arraybackslash}X|>{\centering\arraybackslash}X>{\centering\arraybackslash}X}
\hline
& Method
& \multicolumn{2}{c|}{ScanNet\textsuperscript{*}}
& \multicolumn{2}{c|}{KITTI\textsuperscript{*}}
& \multicolumn{2}{c||}{TartanAir\textsuperscript{*}}
& \multicolumn{2}{c|}{TUM}
& \multicolumn{2}{c}{EuRoC} \\
&
& $t_{\text{rel}}\downarrow$ & $r_{\text{rel}}\downarrow$
& $t_{\text{rel}}\downarrow$ & $r_{\text{rel}}\downarrow$
& $t_{\text{rel}}\downarrow$ & $r_{\text{rel}}\downarrow$
& $t_{\text{rel}}\downarrow$ & $r_{\text{rel}}\downarrow$
& $t_{\text{rel}}\downarrow$ & $r_{\text{rel}}\downarrow$ \\
\hline
\multirow{5}{*}{\rotatebox[origin=c]{90}{\lo Calibrated}} & \lo ORB-SLAM3~\cite{campos2021orbslam3}\textsuperscript{\dag}
& \leavevmode\lo 0.0349 & \leavevmode\lo 3.5730
& \leavevmode\lo 2.5460 & \leavevmode\lo 4.0420
& \leavevmode\lo 0.9261 & \leavevmode\lo 5.2850
& \leavevmode\lo 0.0291 & \leavevmode\lo 4.1360
& \leavevmode\lo 0.0993 & \leavevmode\lo 3.6150 \\
 & \lo LeapVO~\cite{chen2024leapvo}\textsuperscript{\dag}
& \leavevmode\lo 0.0297 & \leavevmode\lo 0.9405
& \leavevmode\lo 2.8890 & \leavevmode\lo 5.0990
& \leavevmode\lo 0.2251 & \leavevmode\lo 0.5736
& \leavevmode\lo 0.0189 & \leavevmode\lo 0.7399
& \leavevmode\lo 0.1645 & \leavevmode\lo 5.8770 \\
 & \lo DPVO~\cite{teed2023dpvo}\textsuperscript{\dag}
& \leavevmode\lo 0.0069 & \leavevmode\lo 0.2090
& \leavevmode\lo 0.1808 & \leavevmode\lo 0.0360
& \leavevmode\lo 0.0213 & \leavevmode\lo 0.0720
& \leavevmode\lo 0.0067 & \leavevmode\lo 0.3920
& \leavevmode\lo 0.0027 & \leavevmode\lo 0.0390 \\
 & \lo iSLAM-VO~\cite{fu2024islam}\textsuperscript{\dag}
& \leavevmode\lo 0.0082 & \leavevmode\lo 0.2350
& \leavevmode\lo 0.1726 & \leavevmode\lo 0.1010
& \leavevmode\lo 0.0992 & \leavevmode\lo 0.1840
& \leavevmode\lo 0.0076 & \leavevmode\lo 0.3720
& \leavevmode\lo 0.0205 & \leavevmode\lo 0.1220 \\
 & \lo MASt3R-SLAM-VO~\cite{murai2024mast3rslam}\textsuperscript{\dag}
& \leavevmode\lo 0.0107 & \leavevmode\lo 0.3995
& \leavevmode\lo 1.1391 & \leavevmode\lo 0.5588
& \leavevmode\lo 0.1062 & \leavevmode\lo 0.4311
& \leavevmode\lo 0.0110 & \leavevmode\lo 0.9577
& \leavevmode\lo 0.0376 & \leavevmode\lo 1.0238 \\
\hline
\multirow{9}{*}{\rotatebox[origin=c]{90}{Uncalibrated}} & ORB-SLAM3~\cite{campos2021orbslam3}\textsuperscript{\dag}
& 0.0398 & 3.7540
& 5.3970 & 3.8130
& 0.8468 & 5.1270
& 0.0300 & 4.5120
& 0.0760 & 2.4320 \\
 & LeapVO~\cite{chen2024leapvo}\textsuperscript{\dag}
& 0.0301 & 0.9801
& 3.2409 & 5.5618
& 0.5059 & 1.6859
& 0.0314 & 1.2760
& 0.1352 & 5.4215 \\
 & DPVO~\cite{teed2023dpvo}\textsuperscript{\dag}
& \underline{0.0082} & 0.2572
& 0.4270 & \textbf{0.0931}
& \underline{0.1065} & 0.2872
& 0.0097 & 0.4907
& 0.0322 & 0.8162 \\
 & iSLAM-VO~\cite{fu2024islam}\textsuperscript{\dag}
& 0.0084 & \underline{0.2426}
& \underline{0.1918} & \underline{0.1104}
& 0.1078 & \textbf{0.2644}
& 0.0087 & \underline{0.3757}
& \underline{0.0277} & \underline{0.5758} \\
 & MASt3R-SLAM-VO~\cite{murai2024mast3rslam}\textsuperscript{\dag}
& 0.0185 & 0.2861
& 1.1595 & 0.5681
& 0.1677 & 0.3076
& 0.0199 & 0.5531
& 0.0558 & 0.6581 \\
 & TSformer~\cite{francani2025tsformer}\textsuperscript{\dag}
& 0.0130 & 0.6660
& 0.2673 & 0.4171
& 0.2873 & 1.6219
& 0.0139 & 1.3621
& 0.0421 & 1.7625 \\
 & MUSt3R~\cite{cabon2025must3r}
& 0.0196 & 0.4896
& 4.7868 & 9.7268
& 0.4594 & 1.4906
& 0.0155 & 0.5137
& 0.1425 & 1.1984 \\
 & AMB3R~\cite{wang2025amb3r}
& 0.0120 & 0.3724
& 0.8941 & 0.2957
& 0.2352 & 1.6842
& \underline{0.0084} & 0.4387
& 0.0439 & 0.9804 \\
 & \ours (Ours)
& \textbf{0.0054} & \textbf{0.1634}
& \textbf{0.1620} & 0.1231
& \textbf{0.1027} & \underline{0.2815}
& \textbf{0.0052} & \textbf{0.3355}
& \textbf{0.0207} & \textbf{0.3133} \\
\hline
\end{tabularx}
\end{table*}

\boldparagraph{Baselines}. No method in the uncalibrated block of \cref{tab:tracking_accuracy}
receives the ground-truth intrinsics. The methods that require intrinsics are instead given a single
GeoCalib~\cite{veicht2024geocalib} estimate from the first frame of each sequence, held fixed for the
run, following the self-calibration protocol of MASt3R-SLAM~\cite{murai2024mast3rslam},
VGGT-SLAM~\cite{maggio2025vggt-slam} and EC3R-SLAM~\cite{hu2025ec3r}; \cref{tab:geocalib} reports how
far those estimates fall from the true values. These are ORB-SLAM3~\cite{campos2021orbslam3},
LeapVO~\cite{chen2024leapvo}, DPVO~\cite{teed2023dpvo} and iSLAM-VO~\cite{fu2024islam}. Their results
with the ground-truth intrinsics are shown in the gray block for reference.

The remaining methods need no intrinsics at all. MUSt3R~\cite{cabon2025must3r},
MASt3R-SLAM-VO~\cite{murai2024mast3rslam} and AMB3R~\cite{wang2025amb3r} recover geometry and camera
pose jointly with a large 3D model, while TSformer~\cite{francani2025tsformer}, like \ours, regresses
relative pose directly from images with no geometric optimization. MASt3R-SLAM-VO optionally accepts
intrinsics, so it also appears in the gray block. Of these, MASt3R-SLAM-VO and TSformer are up to
scale and are $\mathrm{Sim}(3)$-aligned before evaluation, while MUSt3R, AMB3R and \ours predict their
own scale and are evaluated as is.

\boldparagraph{On training data.} The benchmarks marked \textsuperscript{*}\ in
\cref{tab:tracking_accuracy} contribute training data to \ours, with test scenes held out. This is not
a like-for-like advantage, since MUSt3R, MASt3R-SLAM and AMB3R are
built on foundation models pre-trained on substantially more data than our mixture and inherit it
through their backbones. DPVO and iSLAM-VO are trained on TartanAir and
evaluated zero-shot elsewhere, and TSformer is trained on KITTI alone, so KITTI is the only column on
which it is in domain. TUM and EuRoC appear in no method's training data in any form.
The five benchmarks also span a factor of roughly two in normalized focal length, from $f_x/W \approx
0.9$ on ScanNet to $\approx 0.5$ on TartanAir, covering a wide range of camera models.

\subsection{Visual Odometry Results}
As shown in \cref{tab:tracking_accuracy}, \ours attains the best result on all ten calibration-free
columns, and it does so without any test-time optimization. Against the self-calibrated methods as
well, it leads on eight of the ten.
A single set of weights leads across handheld indoor capture on ScanNet and TUM, a micro aerial vehicle
on EuRoC and a car on KITTI, where per-frame translation spans two orders of magnitude. The two columns
it does not win are both rotation columns, where the self-calibrated DPVO~\cite{teed2023dpvo} leads on
KITTI and iSLAM-VO~\cite{fu2024islam} on TartanAir v1. Since $r_{\text{rel}}$ is built from relative rotations, it is invariant to the
$\mathrm{Sim}(3)$ alignment, so the rotation columns already compare every method under identical
conditions.

TSformer~\cite{francani2025tsformer} is the only other method that regresses camera pose directly, so
it is the closest comparison to our architecture and inference scheme. That comparison isolates the two
only on KITTI, where both are in domain and \ours is more accurate on both metrics despite TSformer
receiving a $\mathrm{Sim}(3)$ alignment and \ours none. Its translation error is comparable to the
systems built on large 3D models, so per-frame translation is not what holds direct regression back,
whereas its rotation error is at least three times ours everywhere, a gap that on the four
out-of-domain columns partly reflects training data rather than architecture.

Comparing the two blocks, the two most accurate calibrated methods, DPVO~\cite{teed2023dpvo} and
iSLAM-VO~\cite{fu2024islam}, degrade on all ten columns once they have to calibrate themselves, with
DPVO losing a factor of $2.4$ on KITTI and $12$ on EuRoC translation.
MASt3R-SLAM-VO~\cite{murai2024mast3rslam} degrades on all five translation columns but improves on
four of the five rotation columns. The two weakest calibrated methods,
ORB-SLAM3~\cite{campos2021orbslam3} and LeapVO~\cite{chen2024leapvo}, are inconsistent, and their
completion rates make the comparison harder to read. The errors in \cref{tab:tracking_accuracy} are
averaged over the sequences each method completes. ORB-SLAM3 returns an alignable trajectory for only 55 of the 97
ScanNet scenes, and MASt3R-SLAM-VO fails on 3 of the 18 TartanAir sequences, whereas \ours returns a
trajectory for every sequence of every benchmark.

\boldparagraph{How well is the scale recovered?} Scale is unobservable from monocular geometry, so a
classical pipeline cannot recover it in principle. \ours recovers it from learned priors instead, and
\cref{tab:metric_scale} measures how far that gets. For each test sequence we compute the scale $s$
that the $\mathrm{Sim}(3)$ transformation applies to our trajectory to best align it to the ground
truth, so a median at $s = 1$ means the prediction needs no rescaling. The result is not uniform.
ScanNet, KITTI, and TUM sit within $5\%$ of unity, TartanAir is $15\%$ off, and EuRoC is the weakest
case at $20\%$ below unity, where our trajectories are consistently about a quarter too long. Applying
the alignment nonetheless buys little in per-frame terms, reducing $t_{\text{rel}}$ by at most $12\%$
and making it $23\%$ worse on EuRoC, so scale error and per-frame error are largely independent. The
claim is therefore that \ours recovers scale accurately enough to be evaluated without alignment, not
that it is metric everywhere.

\begin{table}[t]
\centering
\footnotesize
\setlength{\tabcolsep}{4pt}
\renewcommand{\arraystretch}{1.1}
\caption{\textbf{Scale recovered without alignment.} For every test sequence, we compute the scale $s$ of the $\mathrm{Sim}(3)$ transformation that is applied to our trajectory to best align it to the ground truth, so $s = 1$ means no rescaling is needed and $s < 1$ means our trajectory is too long. The up-to-scale baselines of \cref{tab:tracking_accuracy} have no counterpart here, since their scale is supplied by this alignment rather than predicted. We also report $t_{\text{rel}}$ (m/frame) with and without the alignment. $n$ is the number of sequences.}
\label{tab:metric_scale}
\begin{tabular}{lrccc}
\toprule
& & & \multicolumn{2}{c}{$t_{\text{rel}}\downarrow$} \\
\cmidrule(lr){4-5}
Benchmark & $n$ & median $s$ & Aligned & As is \\
\midrule
ScanNet   &  97 & 1.045 & 0.0054 & 0.0054 \\
KITTI     &   2 & 0.996 & 0.1562 & 0.1620 \\
TartanAir &  18 & 1.153 & 0.0950 & 0.1027 \\
TUM       &   9 & 0.952 & 0.0046 & 0.0052 \\
EuRoC     &  11 & 0.796 & 0.0254 & 0.0207 \\
\bottomrule
\end{tabular}
\end{table}

\begin{figure}[!t]
    \centering
    \includegraphics[width=0.85\linewidth]{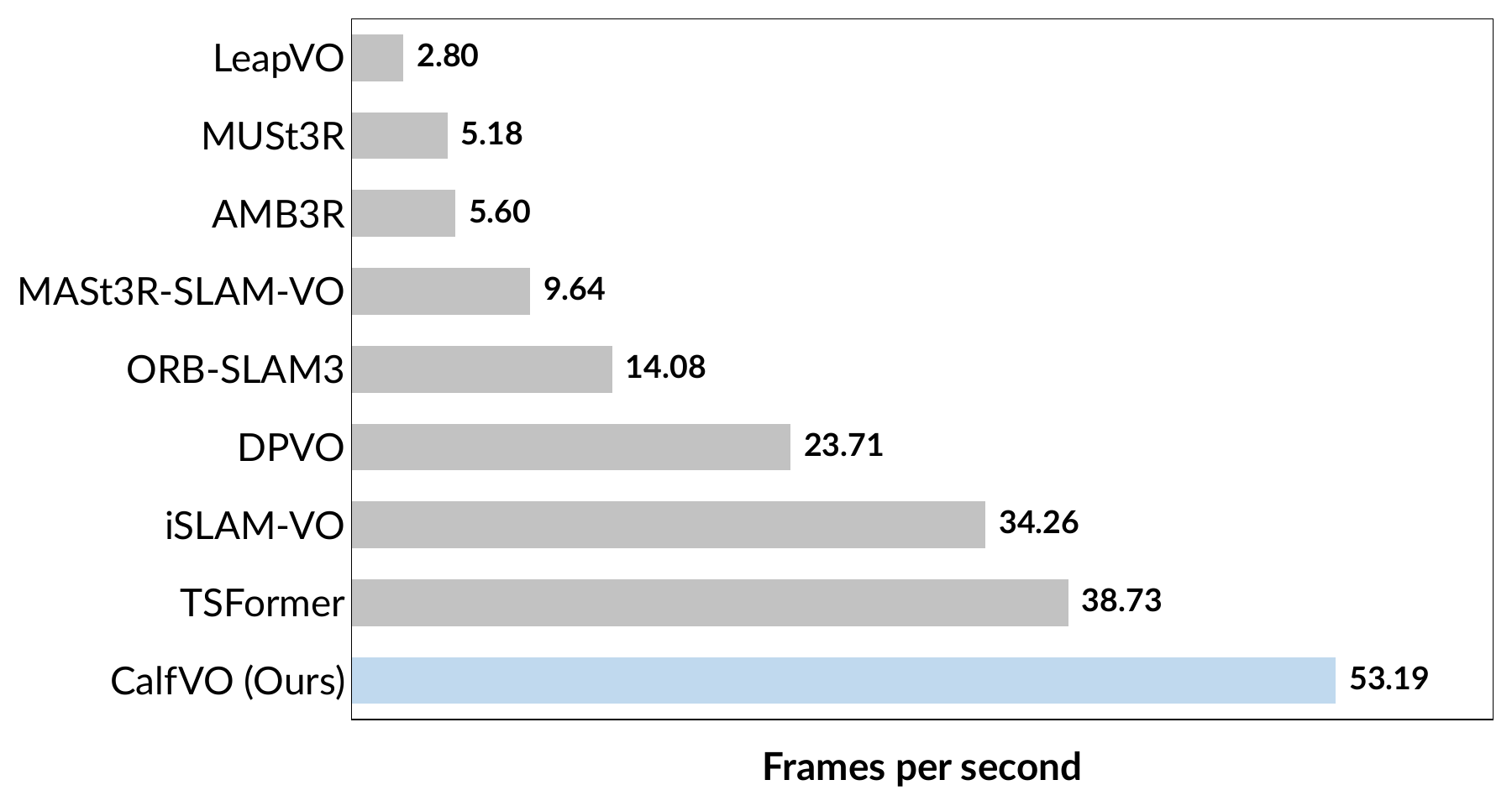}
    \caption{\textbf{Runtime Analysis.} \ours outperforms the baselines in inference speed by avoiding a test-time optimization stage.}
    \label{fig:runtime}
\end{figure}

\boldparagraph{\ours is real-time.} We measure every method on the same 2165 TUM frames,
uncalibrated and at batch size 1, one method at a time on an otherwise idle node with a single NVIDIA
A100 GPU, timing from before the first frame is decoded to after the last pose is produced. \ours is timed at
its shipped inference setting, windows of eight frames overlapping by five. As shown in
\cref{fig:runtime}, \ours runs at 53 FPS and is the fastest method we compare against, by between
$1.6\times$ and $19\times$ against the systems that run an optimization backend and by more than
$9\times$ against the two metric baselines. \ours performs a fixed amount of feed-forward computation
per window, so its cost is linear in the number of frames and independent of the scene, whereas
AMB3R~\cite{wang2025amb3r} pays for a backend whose duration depends on how quickly it converges. The one baseline that also has no backend,
TSformer~\cite{francani2025tsformer}, is the closest at $1.4\times$, which locates the cost in the
backend rather than in the architecture.

\begin{figure*}[!t]
    \centering
    \includegraphics[width=0.88\linewidth]{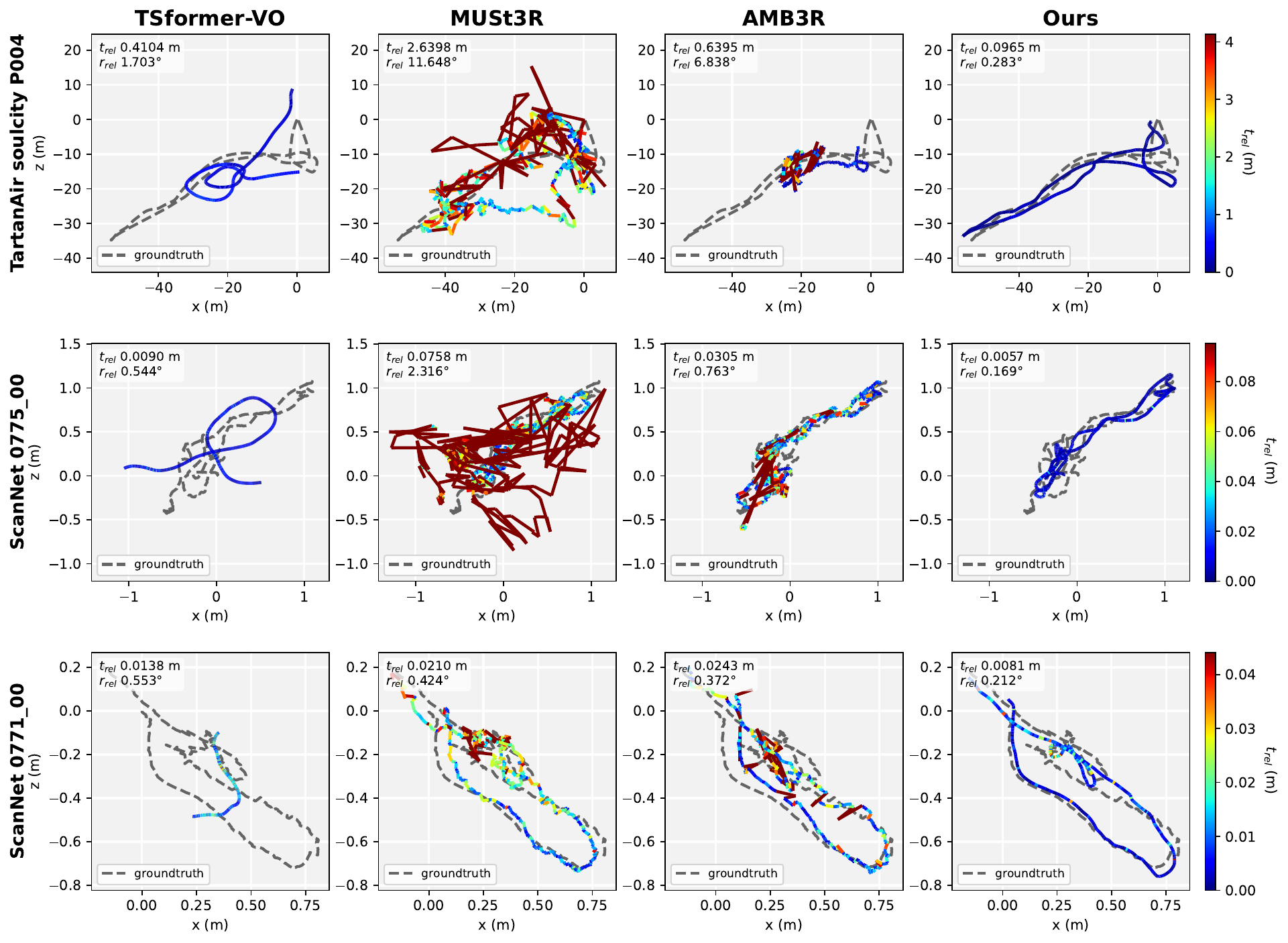}
    \caption{{\bf Trajectory estimation results.} The ground truth is shown as dashed lines. Trajectory color encodes the relative translation error, with higher errors in red and lower in blue.}
    \label{fig:trajetories}
\end{figure*}

\subsection{Ablation Studies}

We ablate four design choices, namely the confidence formulation, the attention factorization, the
rotation parameterization, and the encoder. All four groups of \cref{tab:ablations} share one baseline
run, the gray row, which is at once the with-confidence, time--space, $\mathbb{SO}(3)$ projection and
CroCoV2 configuration. Every other row changes one component and keeps the rest of the final model.
Due to computational limitations, every configuration is trained for 11.4k steps rather than 57k, with
the cosine schedule annealed over that shorter horizon so that each follows a complete schedule. We
report the same metrics as \cref{tab:tracking_accuracy} on the held-out ScanNet~\cite{dai2017scannet}
scenes, so the absolute values are not comparable to it.

%

\begin{table}[t]
\centering
\footnotesize
\setlength{\tabcolsep}{5pt}
\renewcommand{\arraystretch}{1.1}
\caption{\textbf{Ablations.} Relative translation and rotation error on the held-out ScanNet scenes. The \colorbox{gray!20}{gray} row is the shared baseline, run once. \emph{No confidence} is retrained without the confidence formulation, and \emph{Full attention} attends over all tokens of all frames.}
\label{tab:ablations}
\begin{tabular}{lcc}
\toprule
& $t_{\text{rel}}\downarrow$ & $r_{\text{rel}}\downarrow$ \\
\midrule
\multicolumn{3}{l}{\emph{Confidence formulation}} \\
\quad No confidence & 0.0085 & 0.2838 \\
\rowcolor{gray!20} \quad With confidence & \textbf{0.0060} & \textbf{0.1836} \\
\addlinespace[1pt]
\multicolumn{3}{l}{\emph{Attention}} \\
\quad Full attention & 0.0072 & 0.2289 \\
\rowcolor{gray!20} \quad Time--space & \textbf{0.0060} & \textbf{0.1836} \\
\addlinespace[1pt]
\multicolumn{3}{l}{\emph{Rotation parameterization}} \\
\quad Euler angles & 0.0079 & 0.2393 \\
\quad Quaternion & 0.0064 & 0.1874 \\
\quad Gram--Schmidt~\cite{zhou20196d} & 0.0063 & 0.1853 \\
\rowcolor{gray!20} \quad $\mathbb{SO}(3)$ projection & \textbf{0.0060} & \textbf{0.1836} \\
\addlinespace[1pt]
\multicolumn{3}{l}{\emph{Encoder}} \\
\quad DINOv2~\cite{oquab2024dinov2} & 0.0081 & 0.2764 \\
\quad DINOv3~\cite{simeoni2025dinov3} & 0.0079 & 0.2721 \\
\quad $\text{DINOv3}_{\text{VGGT-}\Omega}$ & 0.0079 & 0.2672 \\
\rowcolor{gray!20} \quad $\text{CroCoV2}_{\text{DUSt3R}}$ & \textbf{0.0060} & \textbf{0.1836} \\
\bottomrule
\end{tabular}
\end{table}

\boldparagraph{The gain from the confidences is in the training objective.} \cref{tab:ablations}
compares \ours against a model retrained without the confidence formulation, which regresses pose
directly. The formulation improves $t_{\text{rel}}$ from
0.0085 to 0.0060 and $r_{\text{rel}}$ from 0.2838 to 0.1836, suggesting the effectiveness of the training objective.
Fixed per-term loss weights cannot replace learned log-variances. The loss computes
$w\,\varepsilon\exp(-u) + u$ with $\varepsilon$ the pose error and $u$ a predicted log-variance, so
once $u$ reaches its optimum $u^{\star} = \log(w\varepsilon)$ the gradient on the pose error is
$1/\varepsilon$ regardless of $w$, which leaves only an additive constant $\log w$. Rebalancing rotation against translation therefore requires a structural change
rather than a weight.

\boldparagraph{Time-space attention improves accuracy.} In \cref{tab:ablations}, we replace our
factorized time--space attention with full attention over all tokens of all frames, keeping the number
of blocks and the token dimension fixed. Factorizing attention improves both metrics.

\boldparagraph{SO(3) projection enhances rotation accuracy.} In \cref{tab:ablations}, we evaluate
 four rotation parameterizations, namely Euler angles, quaternions, the Gram--Schmidt
 representation~\cite{zhou20196d} and our projection onto $\mathbb{SO}(3)$. Every predicted
 representation is converted to $\mathbb{SO}(3)$ using~\cite{bregier2021deepregression}, and the same
 geodesic loss is applied in all four cases, so the comparison isolates the parameterization. Our
 projection onto the nearest valid rotation matrix yields the best result. The ordering is the same on both metrics and follows
 how well each parameterization behaves as a map onto $\mathbb{SO}(3)$, with the discontinuous Euler
 angles worst and the double-covering quaternion next. The margin over the Gram--Schmidt
 representation is small, so we report it as a consistent preference rather than a decisive one.

\boldparagraph{Cross-view completion pre-training matters more than pre-training scale.} In
\cref{tab:ablations}, we compare features from four pre-trained encoders, all ViT-Large with roughly
300M parameters, so the comparison separates the pre-training objective from capacity.
DINOv2~\cite{oquab2024dinov2} and DINOv3~\cite{simeoni2025dinov3} are trained for general-purpose
visual representation, whereas $\text{DINOv3}_{\text{VGGT-}\Omega}$~\cite{wang2026vggtomega} and
$\text{CroCoV2}_{\text{DUSt3R}}$~\cite{wang2024dust3r} are trained for 3D geometry. Three of the four
fall within four percent of each other on both metrics, while CroCoV2~\cite{philippe2022croco} trained
within the DUSt3R framework is ahead of all of them by 24 percent on translation and 31 percent on
rotation. $\text{DINOv3}_{\text{VGGT-}\Omega}$ lands with the general-purpose encoders despite
being trained for geometry on substantially more data, so neither a geometric objective in general nor
the scale of pre-training accounts for the gap. We attribute it instead to cross-view completion, which
trains the encoder to establish correspondence across views and so yields features aligned with what
relative pose estimation requires.

\boldparagraph{Limitations and future work.} \ours estimates relative pose over short windows and
integrates the result, so it has no mechanism for correcting global drift. Its error accumulates
monotonically along a sequence, and on long trajectories the absolute error will therefore exceed
that of a system which closes loops or runs a global bundle adjustment, even where the per-frame
error is lower. Our claims are accordingly about relative pose accuracy without calibration, not
about long-horizon global consistency, and combining our predictions with a loop-closure module is a
natural extension. We also do not claim that \ours generalizes to every dataset. Because it is trained
primarily in static environments, its performance may be limited in dynamic settings. Further
improvements could come from more diverse datasets collected across different devices, from systematic
data curation, and from larger pre-trained encoders.

\section{Conclusion}
\label{sec:conclusion}
We presented \ours, a calibration-free approach to monocular visual odometry that regresses relative
camera poses directly from video and aggregates them with confidence-weighted averaging over
overlapping windows. \ours requires no camera intrinsics, no bundle adjustment, and no loop closure, and it recovers scale
from learned priors rather than geometry, accurately enough to be evaluated without any alignment to
the ground truth.

Two of our findings are not specific to this model. A method that never had intrinsics can outperform
ones that lose them, which suggests that it is the dependence on calibration, not the optimization
backend, that limits those systems in the uncalibrated setting. Cross-view completion pre-training
also matters more than the amount of pre-training data, so the accuracy still missing is to be found
in better geometric representations rather than larger ones. Where calibration is unavailable, then,
direct regression is not the weaker option, and what is left to improve is the encoder and the data it
is trained on rather than the geometric optimization we removed.

\ifarxiv 
\boldparagraph{Acknowledgements.} This work was supported by TomTom, the University of Amsterdam, and
the allowance of Top Consortia for Knowledge and Innovation (TKIs) from the Netherlands Ministry of
Economic Affairs and Climate Policy.
\fi

{
    \small
    \bibliographystyle{ieeenat_fullname}
    \bibliography{main}
}

\ifarxiv
  \clearpage
  \appendix
  \setcounter{table}{0}
  \setcounter{figure}{0}
  \setcounter{equation}{0}
  \renewcommand{\thesection}{S\arabic{section}}
  \renewcommand{\thetable}{S\arabic{table}}
  \renewcommand{\thefigure}{S\arabic{figure}}
  \renewcommand{\theequation}{S\arabic{equation}}
\section{Training details}
\label{sec:training}
\boldparagraph{Architecture and optimization.} The encoder is a ViT-Large with 24 blocks, an embedding
dimension of 1024 and a patch size of 16. At an
input resolution of $224 \times 224$ this gives a $14 \times 14$ grid of 196 patch tokens per frame. We train on 16 NVIDIA A100 40\,GB GPUs for roughly 20 hours, which is 57k steps at a measured
1.26\,s per step. The learning rate follows a cosine schedule from $10^{-4}$ to $10^{-5}$ with
1.9k warm-up steps.

\boldparagraph{Data sampling.} Every iteration draws random sequences from the four training streams and a random window within each,
at a frame spacing drawn from the stride range of that stream and a start offset placed anywhere it
fits. TartanAir v2, ScanNet, KITTI and Virtual KITTI contribute 36, 32, 16 and 16 percent of the samples
in each cycle, against the 60, 39, 0.6 and 0.6 percent of a frame-proportional mixture. We oversample
both driving sets because they are the only driving data the model sees. Their stride range is 1 to 2,
against 1 to 5 for the indoor and synthetic sets, since a vehicle advances roughly two orders of
magnitude further between consecutive frames than a handheld camera.

\boldparagraph{Filtering invalid ScanNet poses.} On ScanNet we truncate each sequence at its first invalid pose, which removes roughly 20 percent of the
frames. Jumps above 1.5\,m remain in about one percent of scenes, so we additionally discard any drawn
window whose inter-frame translation exceeds that threshold. No window of a test sequence exceeds it, so
no test trajectory has an interior gap.

\boldparagraph{Translation normalization.} Translations are normalized by a fixed mean and standard deviation that do not match the training
mixture. The mixture standard deviation is set by the driving streams, and normalizing by it compresses
handheld motion below the output resolution of the model. We normalize to the handheld regime instead,
which leaves driving translations several standard deviations out on the forward axis. The L1 translation
loss and the confidence head tolerate this.

\section{Confidence ablation across benchmarks}
\label{sec:conf_all}
\cref{tab:ablations} reports the confidence ablation on held-out ScanNet only. \cref{tab:conf_all}
evaluates the same retrained variant on all five benchmarks.

Removing the confidence formulation is worse on 9 of the 10 columns. Rotation degrades by 37 to 102
percent and translation by 9 to 42 percent. KITTI translation is the only column that does not degrade,
at $-0.3\%$. The gain of the formulation is concentrated in rotation on every benchmark.

\begin{table}[t]
\centering
\footnotesize
\setlength{\tabcolsep}{4pt}
\renewcommand{\arraystretch}{1.1}
\caption{\textbf{Confidence ablation on all five benchmarks.} Relative translation error
$t_{\text{rel}}$ (m/frame) and rotation error $r_{\text{rel}}$ (\degree/frame). Both columns use the
configuration of our final model at a matched $\approx$11.4k steps, so they are not the
\cref{tab:tracking_accuracy} values. $\Delta$ is the relative change of \emph{No conf.} against
\emph{Ours}, so positive is worse.}
\label{tab:conf_all}
\begin{tabular}{llrrr}
\toprule
Benchmark & Metric & Ours & No conf. & $\Delta$ \\
\midrule
\multirow{2}{*}{ScanNet}   & $t_{\text{rel}}$ & \textbf{0.0060} & 0.0085 & $+41.7\%$ \\
                           & $r_{\text{rel}}$ & \textbf{0.1836} & 0.2838 & $+54.6\%$ \\
\addlinespace[1pt]
\multirow{2}{*}{KITTI}     & $t_{\text{rel}}$ & 0.1292 & \textbf{0.1289} & $-0.3\%$ \\
                           & $r_{\text{rel}}$ & \textbf{0.1190} & 0.2321 & $+95.1\%$ \\
\addlinespace[1pt]
\multirow{2}{*}{TartanAir} & $t_{\text{rel}}$ & \textbf{0.1066} & 0.1438 & $+34.9\%$ \\
                           & $r_{\text{rel}}$ & \textbf{0.4137} & 0.8342 & $+101.7\%$ \\
\addlinespace[1pt]
\multirow{2}{*}{TUM}       & $t_{\text{rel}}$ & \textbf{0.0074} & 0.0105 & $+41.6\%$ \\
                           & $r_{\text{rel}}$ & \textbf{0.3462} & 0.5214 & $+50.6\%$ \\
\addlinespace[1pt]
\multirow{2}{*}{EuRoC}     & $t_{\text{rel}}$ & \textbf{0.0217} & 0.0236 & $+8.7\%$ \\
                           & $r_{\text{rel}}$ & \textbf{0.2996} & 0.4096 & $+36.7\%$ \\
\bottomrule
\end{tabular}
\end{table}

\section{Effect of window overlap}
\label{sec:overlap}
The overlap between consecutive windows sets how many estimates each frame pair receives.
\cref{tab:overlap} evaluates the shipped model of \cref{tab:tracking_accuracy} at four overlaps.

Widening the overlap from one to five frames raises the number of estimates per pair from 1 to 2.33 and
reduces $t_{\text{rel}}$ by 1.1 to 6.1 percent and $r_{\text{rel}}$ by 0.8 to 4.2 percent. The largest
gains are on KITTI and TUM. Going to seven frames
of overlap, or 6.97 estimates per pair, gives a further 0.1 to 0.6 percent at $2.3\times$ the evaluation
runtime and is $0.06\%$ worse on TUM rotation. We use five frames. Without any averaging, \ours still has
the lowest $t_{\text{rel}}$ of all uncalibrated methods in \cref{tab:tracking_accuracy} on all five
benchmarks.

\begin{table}[t]
\centering
\footnotesize
\setlength{\tabcolsep}{3pt}
\renewcommand{\arraystretch}{1.1}
\caption{\textbf{Effect of window overlap.} The shipped model evaluated unaligned at four overlaps
between consecutive windows, given as the number of shared frames out of the window size of eight.
An overlap of 5 frames is our setting and is the \ours row of \cref{tab:tracking_accuracy}. The lower
block reports the mean number of estimates per frame pair and the evaluation runtime over all 137
sequences.}
\label{tab:overlap}
\begin{tabular}{llrrrr}
\toprule
& & \multicolumn{4}{c}{Window overlap [frames]} \\
\cmidrule(lr){3-6}
Benchmark & Metric & 1 & 3 & 5 & 7 \\
\midrule
\multirow{2}{*}{ScanNet}   & $t_{\text{rel}}$ & 0.00544 & 0.00539 & 0.00538 & 0.00537 \\
                           & $r_{\text{rel}}$ & 0.16479 & 0.16360 & 0.16340 & 0.16318 \\
\addlinespace[1pt]
\multirow{2}{*}{KITTI}     & $t_{\text{rel}}$ & 0.17256 & 0.16730 & 0.16199 & 0.16144 \\
                           & $r_{\text{rel}}$ & 0.12700 & 0.12345 & 0.12307 & 0.12229 \\
\addlinespace[1pt]
\multirow{2}{*}{TartanAir} & $t_{\text{rel}}$ & 0.10385 & 0.10311 & 0.10273 & 0.10247 \\
                           & $r_{\text{rel}}$ & 0.29389 & 0.28531 & 0.28150 & 0.28030 \\
\addlinespace[1pt]
\multirow{2}{*}{TUM}       & $t_{\text{rel}}$ & 0.00543 & 0.00531 & 0.00524 & 0.00522 \\
                           & $r_{\text{rel}}$ & 0.33930 & 0.33699 & 0.33554 & 0.33574 \\
\addlinespace[1pt]
\multirow{2}{*}{EuRoC}     & $t_{\text{rel}}$ & 0.02103 & 0.02074 & 0.02066 & 0.02063 \\
                           & $r_{\text{rel}}$ & 0.31590 & 0.31418 & 0.31326 & 0.31287 \\
\midrule
\multicolumn{2}{l}{Estimates per pair} & 1.00 & 2.00 & 2.33 & 6.97 \\
\multicolumn{2}{l}{Runtime}           & 5\,min & 6\,min & 9\,min & 21\,min \\
\bottomrule
\end{tabular}
\end{table}

\section{Accuracy of the estimated intrinsics}
\label{sec:geocalib}
The uncalibrated block of \cref{tab:tracking_accuracy} replaces the ground-truth intrinsics of every
method that needs them with a single GeoCalib~\cite{veicht2024geocalib} estimate from the first frame
of each sequence, following the self-calibration protocol of MASt3R-SLAM~\cite{murai2024mast3rslam},
VGGT-SLAM~\cite{maggio2025vggt-slam} and EC3R-SLAM~\cite{hu2025ec3r}. The estimate is constant over a
sequence and therefore acts as a fixed bias rather than as a source of noise. \cref{tab:geocalib}
reports its error.

The principal point is recovered to within half a percent of the image width on ScanNet and exactly on
TartanAir, whose synthetic cameras are perfectly centred. On TUM, KITTI and EuRoC it is off by 11 to 16
pixels, which displaces the optical axis by 0.9 to 1.7 degrees. The focal length is off on average by
10.5 percent on ScanNet, 14 to 28 percent on KITTI, TUM and TartanAir, and 112 percent on EuRoC. Every
benchmark other than KITTI also has a sequence on which the estimate fails outright, with a worst-case
focal error of 66 to 265 percent.

\begin{table}[t]
\centering
\footnotesize
\setlength{\tabcolsep}{3pt}
\renewcommand{\arraystretch}{1.05}
\caption{\textbf{Accuracy of the estimated intrinsics.} Error of the single first-frame
GeoCalib~\cite{veicht2024geocalib} estimate we give the self-calibrated methods in the uncalibrated
block of \cref{tab:tracking_accuracy}, against the ground-truth intrinsics. $|\Delta f|/f$ is the relative focal-length error over the $n$ sequences of each test set.
$\|\Delta c\|$ is the principal-point offset in pixels, and \emph{axis} is the angle it displaces the
optical axis by, $\arctan(\|\Delta c\|/f)$.}
\label{tab:geocalib}
\begin{tabular}{lrrrrrr}
\toprule
& & \multicolumn{3}{c}{$|\Delta f| / f$} & \multicolumn{2}{c}{Principal point} \\
\cmidrule(lr){3-5}\cmidrule(lr){6-7}
Benchmark & $n$ & mean & med. & max & $\|\Delta c\|$ & axis \\
\midrule
ScanNet   & 97 & 10.5\% & 7.5\%  & 74.4\%  & 2.7\,px  & 0.27\degree \\
KITTI     &  2 & 14.1\% & 14.1\% & 16.0\%  & 11.3\,px & 0.91\degree \\
TartanAir & 18 & 27.6\% & 19.6\% & 87.1\%  & 0.0\,px  & 0.00\degree \\
TUM       &  9 & 22.6\% & 19.8\% & 66.2\%  & 15.4\,px & 1.70\degree \\
EuRoC     & 11 & 112.4\% & 87.1\% & 264.8\% & 12.1\,px & 1.52\degree \\
\bottomrule
\end{tabular}
\end{table}

\section{Decoder attention maps}
\label{sec:attention_maps}
\Cref{fig:attn_maps} shows where the decoder attends when it predicts a relative pose. The attention
concentrates on the image regions that correspond to the query rather than spreading over the frame. The
decoder therefore recovers a correspondence-like signal from pose supervision alone, without the
matching loss or the keypoint detector that classical odometry relies on.

\begin{figure*}[!t]
    \centering
    \includegraphics[width=\linewidth]{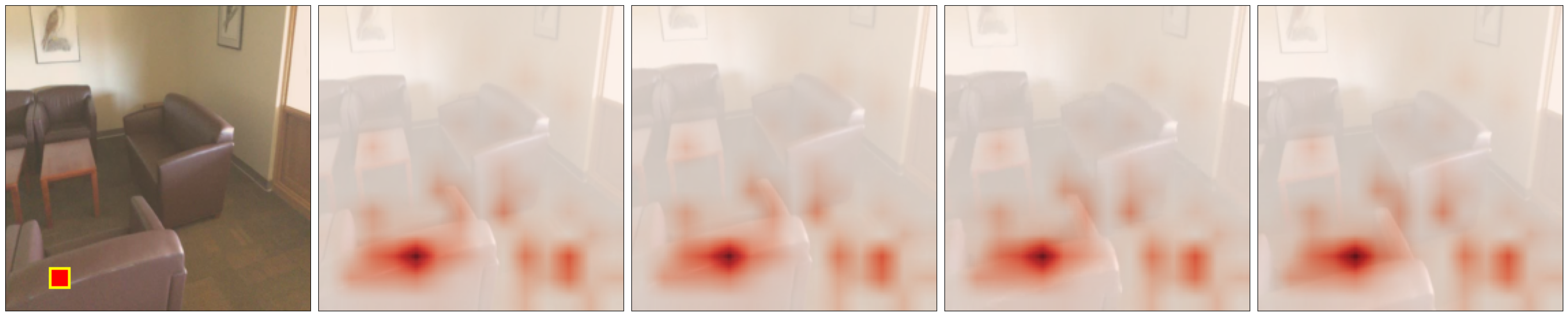}
    \includegraphics[width=\linewidth]{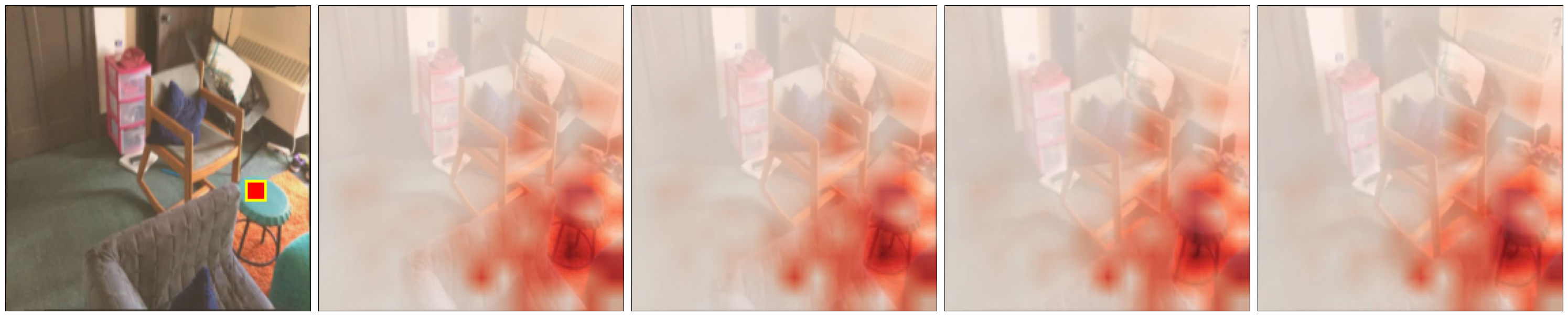}
    \includegraphics[width=\linewidth]{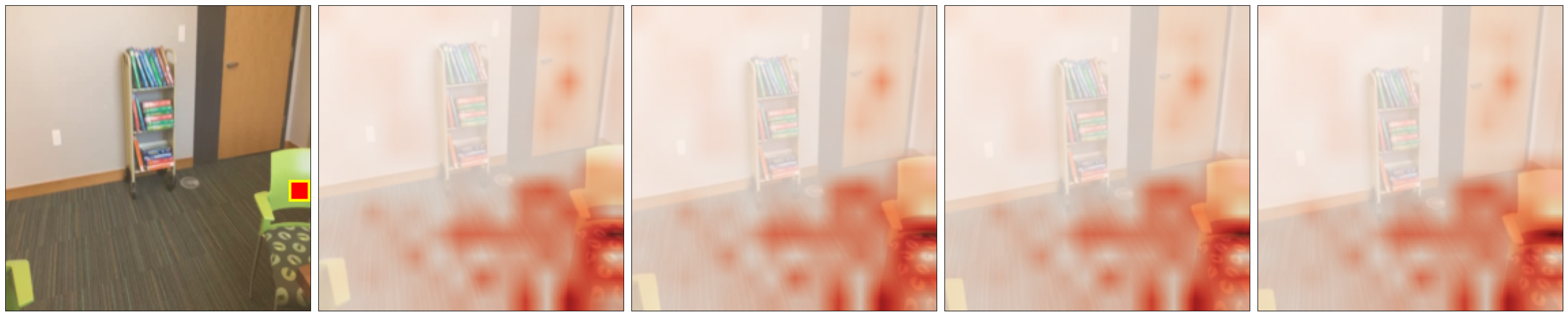}
    \caption{{\bf Attention maps from the \ours decoder.} Each row shows an original image with a selected query (red square), followed by the attention that query receives in the four subsequent frames of the same window.}
    \label{fig:attn_maps}
\end{figure*}

\section{Sequence completion}
\label{sec:completion}
The errors in \cref{tab:tracking_accuracy} are averaged over the sequences for which each method
returns a trajectory, and that subset varies between methods. ORB-SLAM3~\cite{campos2021orbslam3} is
affected the most. Only 55 of the 97 ScanNet sequences yield an estimate that can be aligned to the
ground truth with the Umeyama algorithm~\cite{umeyama1991lse}, so its ScanNet error is measured on
little more than half of the test set. MASt3R-SLAM-VO~\cite{murai2024mast3rslam} fails on 3 of the 18
TartanAir v1 sequences in the uncalibrated setting and on 7 of 18 when it is given the ground-truth
intrinsics. \ours returns a trajectory for every sequence of every benchmark, since it performs a fixed
amount of feed-forward computation per window and has no initialization or optimization stage that can
diverge.

\section{Failure mode analysis}
\label{sec:failure}
Odometry estimates the motion between frames, and a trajectory is the composition of those relative
estimates. In a SLAM system a backend keeps that composition globally consistent through loop closure
and global bundle adjustment, which \ours does not perform. The relative errors of
\cref{tab:tracking_accuracy} therefore measure the quantity our model predicts, whereas the absolute
trajectory error and absolute rotation error of \cref{tab:failure} measure the composition as well.

\begin{table}[t]
    \centering
    \footnotesize
    \setlength{\tabcolsep}{5pt}
    \renewcommand{\arraystretch}{1.15}
    \caption{\textbf{Global error of \ours.} Absolute trajectory error (m) and absolute rotation error (\degree) over whole trajectories, evaluated as is and after the $\mathrm{Sim}(3)$ alignment. Compare with the relative errors of \cref{tab:tracking_accuracy}, which are computed from the same predictions.}
    \label{tab:failure}
    \begin{tabular}{lcccc}
    \toprule
    & \multicolumn{2}{c}{ATE {[m]}$\downarrow$} & \multicolumn{2}{c}{ARE {[\degree]}$\downarrow$} \\
    \cmidrule(lr){2-3} \cmidrule(lr){4-5}
    Benchmark & As is & Aligned & As is & Aligned \\
    \midrule
    ScanNet   & 0.431  & 0.187  & 8.400  & 10.377 \\
    KITTI     & 93.430 & 23.275 & 15.054 & 10.850 \\
    TartanAir & 10.654 & 4.447  & 24.358 & 20.343 \\
    TUM       & 0.372  & 0.154  & 8.900  & 9.482  \\
    EuRoC     & 3.968  & 2.249  & 55.647 & 63.861 \\
\bottomrule
    \end{tabular}
\end{table}

The absolute error tracks the spatial extent of a trajectory rather than the per-frame error.
ScanNet and TUM stay below half a metre, while KITTI reaches 93\,m even though its relative
translation error is the best of any uncalibrated method in \cref{tab:tracking_accuracy}. This is the
cost of removing the optimization backend.

The $\mathrm{Sim}(3)$ alignment reduces the absolute trajectory error by a factor of 1.8 to 4.0. On
ScanNet, KITTI and TUM the median per-sequence scale factor is within $5\%$ of unity
(\cref{tab:metric_scale}), so almost none of that reduction comes from rescaling and what the alignment
removes is accumulated orientation error. Rotation drift rather than scale is the dominant term in our
global error on those three benchmarks. On TartanAir and EuRoC, whose median scale factors are 1.153 and
0.796, both terms contribute.

The alignment does not consistently improve the absolute rotation error, which grows on ScanNet, TUM
and EuRoC and falls on TartanAir and KITTI. The $\mathrm{Sim}(3)$ fit minimizes positional error, so it
accepts a worse orientation where a small rotation buys a large positional gain. The absolute rotation
error of an aligned trajectory is therefore not comparable across methods.

One limitation is not visible in these tables. Our training data is almost entirely static, so we
expect degradation on scenes with substantial independent motion, which none of our five benchmarks
contains in quantity.

\fi

\end{document}